\documentclass[letterpaper, 10 pt, conference]{ieeeconf}
\IEEEoverridecommandlockouts                              

\usepackage{times}
\usepackage{xcolor}
\usepackage{dsfont}
\usepackage[font={small}]{caption}

\usepackage{multicol}
\usepackage[bookmarks=true]{hyperref}
\usepackage{amssymb}
\usepackage{graphicx}
\usepackage{amsmath}
\usepackage{algorithm}
\usepackage{algpseudocode}
\newtheorem{prob}{Problem}

\usepackage{tikz}
\usepackage{subcaption}
\usepackage{bm}
\usepackage{caption} 
\usepackage{makecell}
\usepackage{booktabs}
\usepackage{cite}

\begin{document}

\title{\LARGE \bf
DiffCoTune: Differentiable Co-Tuning for Cross-domain Robot Control
}




\author{ 
Lokesh Krishna, 
Sheng Cheng, 
Junheng Li, 
Naira Hovakimyan, 
Quan Nguyen
\thanks{L. Krishna, J. Li and Q. Nguyen are with the  Department of Aerospace and Mechanical Engineering, USC Viterbi School of Engineering, CA, USA.}
\thanks{S. Cheng and N. Hovakimyan are with the Department of Mechanical Science and Engineering, University of Illinois Urbana-Champaign, IL, USA.}
}

\maketitle

\begin{abstract}
The deployment of robot controllers is hindered by modeling discrepancies due to necessary simplifications for computational tractability or inaccuracies in data-generating simulators. Such discrepancies typically require ad-hoc tuning to meet the desired performance, thereby ensuring successful transfer to a target domain. We propose a framework for automated, gradient-based tuning to enhance performance in the deployment domain by leveraging differentiable simulators. Our method collects rollouts in an iterative manner to co-tune the simulator and controller parameters, enabling systematic transfer within a few trials in the deployment domain. Specifically, we formulate multi-step objectives for tuning and employ alternating optimization to effectively adapt the controller to the deployment domain. The scalability of our framework is demonstrated by co-tuning model-based and learning-based controllers of arbitrary complexity for tasks ranging from low-dimensional cart-pole stabilization to high-dimensional quadruped and biped tracking, showing performance improvements across different deployment domains. 

\end{abstract}

\IEEEpeerreviewmaketitle


\section{Introduction}
\label{sec:intro}

Meeting a desired performance while transferring robot controllers from a \emph{synthesis domain}, like a simulator or dynamics model, to a \emph{deployment domain}, like the real system, remains a ubiquitous challenge across synthesis paradigms due to uncertainties from unmodeled dynamics, noise, and perturbations. Model-based control synthesis offers well-defined techniques to address uncertainty, namely robust or stochastic control \cite{ahmad2020stochastic}, adaptive control \cite{mohsen2024adaptive}, and system identification (sysId)~\cite{atkeson1986estimation}. In principle,  \emph{robust or stochastic} control designs consider the worst case or expectation over some distribution of the uncertainty; \emph{adaptive} control dynamically adjusts parameters based on an estimation of uncertainty; and \emph{system identification} aims to tune the synthesis domain's parameters to best reflect the deployment domain using a physics- or first-principle-based model. Scaling these distinct notions, model-free learning-based control has conceptually equivalent techniques, namely domain randomization ~\cite{peng2018sim}, domain adaptation \cite{peng2020learning}, and sysId using deep neural networks (DNNs)~\cite{jemin2019learning}. Domain randomization assumes an uncertainty distribution like stochastic control, while domain adaptation estimates invariant features or parameters like adaptive control. Each method, when applied individually, has specific limitations: robust control and domain randomization can yield conservative performance under a poor choice of uncertainty bounds \cite{tan2018sim,tiboni2024domain}, adaptive control and domain adaptation require estimateable uncertainty from the observable states~\cite{yu2020learning}. To this end, given a 
reasonable robot model or simulator
a combination of these methods has been employed to achieve empirical success, such as robust adaptive control \cite{petros1995robust,naira2010l1}, domain adaptation with randomization \cite{peng2020learning,kumar2021rma}, and domain randomization with DNN-based sysId \cite{jemin2019learning}.
\begin{figure}[t]
    \centering
    \includegraphics[clip, trim=0.6cm 0.4cm 0.7cm 0.3cm, width=1\columnwidth]{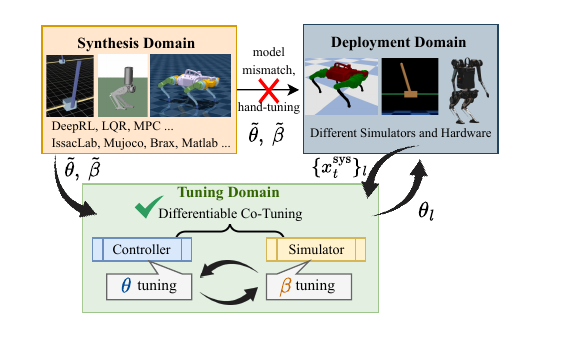}
    \caption{Overview of the proposed automated co-tuning approach for cross-domain transfer. 
    Accompanying video results at : \href{https://youtu.be/eRRVczgZAy0}{\color{blue}link}
    }
    \label{fig:ov}
    \vspace{-0.8cm}
\end{figure}
In a similar vein, we focus on the intersection of domain adaptation and system identification for the automated transfer of robot controllers. Starting with a nominal controller designed in the synthesis domain, we aim to adapt it for the deployment domain by simultaneously calibrating model parameters in an intermediate tuning domain. Thus, we formulate the cross-domain transfer as a co-tuning problem and develop a scalable technique for solving it across different systems and controllers.

To this end, we propose to study system identification and controller adaptation as a coupled problem, with the sole focus being performance improvement in the deployment domain. Given a nominal differentiable controller, we leverage a differentiable simulator to co-tune the parameters of the controller and the simulator, utilizing a handful of rollouts in the deployment domain. Thus, our proposed formulation eases the demand on a global optimum (as usually required in sysId or control synthesis) by requiring neither an accurate recreation of the target system nor the synthesis of a controller from scratch. Instead, we only seek the adaptation of the nominal controller to perform better in the deployment domain.
Such a perspective aims to tune existing empirically compelling model-based \cite{li2024continuous} and model-free~\cite{krishna2024ogmp} controllers, utilizing gradient-based updates for searching an improved local solution within the neighborhood of the nominal controller.

Recent approaches employ \emph{on-policy} reinforcement learning (RL) to learn residual policies compensating for unmodelled dynamics, followed by fine-tuning the nominal policy in the augmented simulator \cite{he2025asapaligningsimulationrealworld, fey2025bridgingsimtorealgapathletic}. Such multi-stage optimization for tuning relies on reward shaping and has only been shown on specific systems with learnt controllers. In contrast, we aim to develop a generic tuning framework without the need to tune its own set of hyperparameters. 
Our proposed iterative approach is similar in principle to \cite{levy2024learning} alternating between gradient-based optimization to refine the differentiable simulator from the rollouts and model-free Deep RL for learning a policy in the updated simulator from scratch. While \cite{levy2024learning} relies on a carefully engineered choice of low-level gait controller specific to the quadrupedal locomotion, we focus on pure gradient-based co-tuning of a nominal controller on systems of arbitrary complexity without task-specific design requirements.
The main contribution of this work is twofold:          
\begin{enumerate}
    \item We present co-tuning techniques to simultaneously tune a nominal controller with a differentiable simulator for an automated and systematic cross-domain transfer, improving performance with a few trials ($<5$) in the deployment domain. 
    \item Experimental evaluation spans systems from low-dimensional cart-pole stabilization to high-dimensional quadruped and bipedal tracking. The proposed approach improves the performance of diverse controllers such as LQR, PD, MPC, and DNN policies, successfully transferring to different simulators and hardware.  
\end{enumerate}

\section{Related work}
\noindent \textbf{Controller auto-tuning:            }
Previous auto-tuning methods can be categorized into model-free~\cite{marco2016automatic, berkenkamp2016safe,calandra2014bayesian,lizotte2007automatic,loquercio2022autotune, song2022policy,GIBO} and model-based~\cite{trimpe2014self,kumar2021diffloop,cheng2024diff,cheng2023difftunePlus,tao2024difftunempc}. 
Both approaches iteratively pick the parameter candidate for evaluation that is likely to improve the performance over the previous trials. 
Model-free auto-tuning methods 
apply a data-driven model to capture the relationship between a parameter choice and the performance metric.
Representative approaches include Markov chain Monte Carlo~\cite{loquercio2022autotune}, policy search~\cite{song2022policy}, and Bayesian optimization~\cite{marco2016automatic, berkenkamp2016safe,calandra2014bayesian,lizotte2007automatic,moriconi2020high,GIBO}. 
Such approaches do not rely on the knowledge of the model and controller and are compatible with physical systems' data owing to their data-driven nature.
However, the absence of physics or first-principle model knowledge (offering a low-dimensional compact representation to the system) is remedied by data-driven models (DNN or Gaussian processes) that often yield higher-dimensional parameter space to characterize the system. In particular, 
Bayesian optimization is hardly scalable to tuning in high-dimensional ($>$20) parameter spaces~\cite{berkenkamp2016safe}. 
Model-based auto-tuning methods leverage the knowledge of the system model and a model-based controller, often applying gradient descent so that the performance can improve based on the local gradient information~\cite{trimpe2014self,kumar2021diffloop,cheng2024diff,cheng2023difftunePlus,tao2024difftunempc}.
However, model-based auto-tuning suffers from biased gradients due to model mismatch that originated from uncertainties or time-varying system configurations, which is one cause of the sim-to-real gap from the perspective of controller tuning. Previous work~\cite{cheng2024diff} has applied the $\mathcal{L}_1$ adaptive control~\cite{wu2025L1Quad} to enforce the system to behave closely to a nominal uncertainty-free system and to preserve the gradient from being biased. However, the nominal model itself can be tuned in the process to reach the same goal, which will be explored in this paper.

\noindent \textbf{Differentiable simulators:}
The recent progress in Auto Differentiation (AD) tools~\cite{jax2018github} and, thereby, differentiable simulators \cite{brax2021github,howell2022dojo} have made it possible to scale gradient computation for high-dimensional systems \cite{brax2021github,howell2022dojo}. Prior works have leveraged differentiable simulators in the context of either system identification \cite{eric2021neuralsim, le2021differntiable, murthy2021gradsim} with the focus on physical realism or a nominal control synthesis \cite{song2024learning,luo2024residual,nina2023training} in a fixed ideal model. While differentiable simulation-based system identification may aid in synthesizing more transferable controllers, it depends on the quality of the identified model. Furthermore, the model suits the data collected from a particular task, leaving it a challenging mission for the same model to facilitate transfer to a different task (e.g., walking vs. parkour for a quadruped).
\section{Problem Formulation}
\label{sec:formulation}

A simulator represented as the discrete-time dynamics model
\begin{equation}
\label{eq:sim}
    x_{t+1} = f(x_t, u_t ; \beta),
\end{equation} 
which is 
paramterized by $\beta$, and $x_0$ is known. Note that $\beta$ could be the physical parameters like mass and friction, as well as augmented terms like parameters of the residual dynamics neural network. Thus, Eq. \eqref{eq:sim} is assumed to be a sufficiently parameterized model to fit the real-world system to arbitrary accuracy. Eq. \eqref{eq:sim} is called a \emph{differntiable simulator} if it provides $\nabla_{x_t}f$, $\nabla_{u_t} f$ and $\nabla_{\beta} f$. For the remainder of the paper, we refer to Eq. \eqref{eq:sim} as the model and refer to the target system as sys.  
The model belongs to the tuning domain, and the sys belongs to the deployment domain, as visualized in Fig. \ref{fig:ov}. 

A controller or a state-feedback policy is represented as
\begin{equation}
    u_t = \pi(x_t ; \theta),
\end{equation}
which is paramterized by $\theta$. We can readily extend to output-feedback policies of the type  $ u_t = \pi(y_t ; \theta)$ with a differentiable state to output map $y_t=g(x_t)$. A \emph{differntiable controller} is one where $\nabla_{x_t}\pi$ and $\nabla_{\theta}\pi$ are computable.

A \emph{system-identification objective} for the model Eq. \ref{eq:sim} over a trajectory of length $T$ is given as
\begin{equation}
 \label{eq:sysid}
   J^\text{sysId}(\theta,\beta, \{x^\text{sys}_t\}) := \frac{1}{T}\sum_{t=1}^{T} \|x_t - x^\text{sys}_t \|^2_2,
\end{equation}
where $x^\text{sys}_t$ is the observable state of the target system whose elements are either directly measured or estimated. Thus, we can think of the system as a map from control input to such observable states: $x^\text{sys}_{t+1}=\text{sys}(u_t)$. We denote a finite state trajectory unrolled with $u_t=\pi(x_t;\theta)$ by  $\{x^\text{sys}_t\}$. For a model $f$ defined on this state space, rolling out the same controller results in: $x_{t+1} = f(x_t, \pi(x_t;\theta); \beta)$. While the system map sys$(\cdot)$ is unknown, acting as a black box permitting inference only, the model's dynamics function $f$ is typically well-defined and computationally transparent.   
Note that our definition of $J^\text{sysId}$ compares the closed-loop dynamics of the model with the system for a given controller. In conventional sysId, the open-loop dynamics of the model over a sequence of controller-agnostic control inputs is compared with the corresponding system response.  

A \emph{task objective} for the controller is given as 
\begin{equation}
 \label{eq:perf_real}
    J^\text{task}(\theta,\text{sys}) := \frac{1}{T} \sum_{t=1}^{T} c(x^\text{sys}_t, u_t), 
\end{equation}
where $x^\text{sys}_t$ are states rolling-out $x^\text{sys}_{t+1} = \text{sys}(\pi(x^\text{sys}_t; \theta))$ in the target-system and $c(\cdot)$ is a differentiable loss term quantifying the performance, which shall be minimized. Given the unknown map sys$(\cdot)$, a differentiable optimization of Eq. \ref{eq:perf_real} is not possible. Hence, consider the following task-objective defined on simulated rollouts,   
\begin{equation}
 \label{eq:perf_sim}
    J^\text{task}(\theta,\beta) := \frac{1}{T}\sum_{t=1}^{T} c(x_t, u_t),
\end{equation}
where $x_t$ is the state by rolling-out the controller $\pi(\cdot;\theta)$ in the model $f$, i.e., $x_{t+1} = f(x_t, \pi(x_t;\theta); \beta)$. Note that Eq. \ref{eq:perf_sim} provides additional decision variables $\beta$ to improve the performance alongside $\theta$. 
\begin{prob}
\label{prb:main}
Given an initial guess of the model parameters $\tilde{\beta}$ and a correspondingly synthesized nominal controller with parameter $\tilde{\theta}$, the \emph{tuning problem} is to obtain a fine-tuned controller for the system, denoted by $\theta^*$, such that 
\begin{equation}
\label{eq:prob1}
    J^\text{task} ( \theta^* , \text{sys}) \leq J^\text{task} (\tilde{\theta}, \text{sys}). 
\end{equation}
\end{prob}

Since $J^\text{task} (\cdot, \text{sys})$ is non-differentiable, we aim to leverage $J^\text{task} (\cdot, \beta)$ in Eq.~\eqref{eq:perf_sim} as a differentiable surrogate. 
Utilizing $J^\text{task} (\cdot, \beta)$ to solve Problem \ref{prb:main} poses a new challenge: finding feasible $\beta$ for improving the task performance in the target system. Specifically, an update in $\beta$ should 
facilitate a performance improvement.  

\section{Proposed Approach}
\label{sec:approach}



Assuming access to a ``handful" of rollouts in the deployment domain, we are interested in an iterative tuning algorithm to co-tune the model and controller parameters for improving performance. Given an initial point ($\tilde{\beta}, \tilde{\theta}$), in each tuning iteration, we assume access to do a \emph{single rollout}, $\{x^\text{sys}_t\}_l$ in the target system and use it to simultaneously update for $\beta$ and $\theta$ using a ``UpdateStrategy". With a maximum number of $L$ iterations on the target system, we thus search for a $\theta^*$ solving Problem \ref{prb:main},  as shown in Alg. \ref{alg:offline}. In every tuning iteration, $\beta_{l}$ is expected to be updated along a direction feasible for performance improvement in the target system. To this end,  minimizing $J^\text{sysId}$ drives the closed-loop dynamics of the model closer to the target system, thereby making it a requirement for such a feasible $\beta$. Thus, if $f$ and $\pi$ are differentiable, a gradient-descent approach can be used on some combination of $J^\text{task}$ and $J^\text{sysId}$ to simultaneously tune the model as well as the controller for system performance. Having outlined the high-level tuning algorithm in Alg. \ref{alg:offline}, we now discuss two different formulations for the UpdateStrategy: \emph{Combined} and \emph{Split-Alternate}. 
\begin{algorithm}
\caption{Iterative Co-Tuning}
\label{alg:offline}
\begin{algorithmic}[1]

\Require $(\tilde{\beta}, \tilde{\theta})$, $x_0$, $L$, UpdateStrategy()
\State \textbf{intialize} $\beta_0 = \tilde{\beta},\, \theta_0 = \theta_\text{best} = \tilde{\theta}$
\For{$l = 0$ to $L$}
\State Rollout $\pi(\,.\,;\theta_l)$ from $x_0$ in the system to collect $\{x^\text{sys}_t\}$
\State $\theta_{l+1}, \beta_{l+1} \leftarrow$ UpdateStrategy($\beta_l, \theta_l, \{x^\text{sys}_t\}) $ 
\If{ $J^\text{task}(\theta_\text{best},\text{sys}) \geq  J^\text{task}(\theta_{l+1},\text{sys})$ }  
    \State $\theta_\text{best} = \theta_{l+1}$
\EndIf
\EndFor \\
\Return $\theta^* \leftarrow \theta_{\text{best}}$
\end{algorithmic}
\end{algorithm}

\subsection{Update Strategy: Combined} 
\label{subsubsec:comb}A straightforward formulation would be to optimize the weighted sum of the two objectives of choice as follows,
\begin{align}
    \label{eq:comb_prob}
    \begin{bmatrix}
        \beta_{l+1}\\ \theta_{l+1}
    \end{bmatrix} = \arg & \min_{[\theta,\beta]^T} \,\, \underbrace{ w^\text{task}J^\text{task}(\theta,\beta) + w^\text{sysId}J^\text{sysId}(\theta,\beta, \{x^\text{sys}_t\}_l)}_{J^\text{comb}}  \\
            \text{s.t.} & \quad x_{t+1} = f(x_t, u_t ; \beta),\, u_{t} = \pi(x_t; \theta), \nonumber 
\end{align}
where $w^\text{task}$ and $w^\text{sysId}$ are corresponding weights. We numerically solve Eq. \ref{eq:comb_prob} with a maximum number of epochs $K$ with a termination condition $\mathcal{C}$  as shown in Alg. \ref{alg:usc}.

    


\begin{table*}[t]
  \centering
    \caption{Overview of the considered system, task, and controller choices}
  \begin{tabular}{ccccccccc}
    \toprule
    system&
    $\pi$ type & 
    synthesis domain&
    dim($x$)& 
    dim($u$)& 
    dim($\theta$)&
    $\beta$ &
    task &
    deployment domain \\
    \midrule
    cartpole&
    $\theta^Tx$ & 
    LQR w/ linearized model&
    4 & 
    1 & 
    4 &
    \makecell{link masses, gear ratio \\  joint friction} &
    stabilize &
    MuJoCo \\
    
    cartpole&
    DNN($x$;$\theta$) & 
    PPO w/ IssacLab&
    4 & 
    1 & 
    1249 &
    link masses &
    stabilize &
    MuJoCo \\

    biped&
    QP($g(x)$;$\theta$) & 
    MPC w/ SRBM&
    32 & 
    10 & 
    3 &
    link masses &
    track &
    MuJoCo \\

    biped&
    PD($g(x)$;$\theta$) & 
    TO w/ varying models&
    32 & 
    10 & 
    1010 &
    link masses &
    track &
    MuJoCo, Hardware \\

    quadruped&
    DNN($g(x)$;$\theta$) & 
    PPO w/ Brax&
    36 & 
    12 & 
    112280 &
    link masses &
    track &
    Pybullet, Hardware \\
    \bottomrule
    
    \hline
  \end{tabular}

\label{table:stc}
  \vspace{-4mm}
\end{table*}

\begin{algorithm}[h]
\caption{UpdateStrategy: Combined}
\label{alg:usc}
\begin{algorithmic}[1]
\Require $(\beta_l, \theta_l)$, $x_0$, $K$, $\{x^\text{sys}_t\}_l$
\State \textbf{intialize} $\beta_0 = \beta_l,\, \theta_0 = \theta_l$
\For{$k = 0$ to $K$ \textbf{while} $\mathcal{C}$ is false}
\State Rollout $\pi(\,.\,;\theta_k)$ from $x_0$ in $x_{t+1} = f(x_t, u_t ; \beta_k)$ 
\State Compute $\nabla_{\beta_k,\theta_k} J^\text{comb}$
\State $\left[ \begin{matrix} \theta_{k+1} \\ \beta_{k+1} \end{matrix} \right] = \left[ \begin{matrix} \theta_{k} \\ \beta_{k} \end{matrix} \right] - \text{GradientUpdate}(\nabla_{\beta_k,\theta_k} J^\text{comb})$ 
\EndFor \\
\Return $(\beta_{l+1}, \theta_{l+1}) \leftarrow (\beta_k, \theta_k) $
\end{algorithmic}
\end{algorithm}
\subsection{Update Strategy: Split-Alternate} Given two qualitatively distinct parameter sets $\theta$ and $\beta$ and the structure of the objective functions, we propose the following alternating optimization\cite{bezdek2002some} based  formulation
\begin{align}
    \label{eq:split_prob_beta}
    \beta_{l+1} = \arg & \min_{\beta} \,\, J^\text{sysId}(\theta_l,\beta,\{x^\text{sys}_t\}_l) \\
            \text{s.t.} & \quad x_{t+1} = f(x_t, u_t ; \beta),\, u_{t} = \pi(x_t; \theta_l), \nonumber \\
    \label{eq:split_prob_theta}
    \theta_{l+1} = \arg & \min_{\theta} \,\, J^\text{task}(\theta,\beta_{l+1}) \\
            \text{s.t.} & \quad x_{t+1} = f(x_t, u_t ; \beta_{l+1}),\, u_{t} = \pi(x_t; \theta) .\nonumber 
\end{align}

Equations  \eqref{eq:split_prob_beta} and \eqref{eq:split_prob_theta} split the update into two sub-problems and solve them sequentially, first for $\beta$ and then $\theta$. While solving Eq. \eqref{eq:split_prob_beta}, $\theta$ is fixed and alternatively for Eq. \eqref{eq:split_prob_theta}, $\beta$ is fixed. Intuitively, in each update, using $\{x^\text{sys}_t\}_l$, we first update the simulator for improved fidelity and then tune the controller in the updated simulator as outlined in Alg. \ref{alg:usa} with identical design choices (learning rate and maximum epochs, $K$) as the combined variant. To keep the number of gradient updates consistent with the combined variant, we allocate a maximum of $\frac{K}{2}$ epochs for both the $\beta$ and $\theta$ updates. 

\begin{algorithm}[h]
\caption{Update Strategy: Split-Alternate}
\label{alg:usa}
\begin{algorithmic}[1]
\Require $(\beta_l, \theta_l)$, $x_0$, $K$, $\{x^\text{sys}_t\}$
\State \textbf{intialize} $\beta_0 = \beta_l,\, \theta_0 = \theta_l$
\For{$k = 0$ to $\frac{K}{2}$ \textbf{while} $\mathcal{C}$ is false}
\State Rollout $\pi(.;\theta_l)$ from $x_0$ in $x_{t+1} = f(x_t, u_t ; \beta_k)$
\State Compute $\nabla_{\beta_k} J^\text{sysId}$ along the rollout
\State $\beta_{k+1} = \beta_{k}  - \text{GradientUpdate}(\nabla_{\beta_k} J^\text{sysId})$ 
\EndFor
\For{$i = 0$ to $\frac{K}{2}$ \textbf{while} $\mathcal{C}$ is false}
\State Rollout $\pi(.;\theta_i)$ from $x_0$ in $x_{t+1} = f(x_t, u_t ; \beta_k)$
\State Compute $\nabla_{\theta_i} J^\text{task}$ along the  rollout
\State $\theta_{i+1} = \theta_{i}  - \text{GradientUpdate}(\nabla_{\theta_i} J^\text{task}$) 
\EndFor \\
\Return $(\beta_{l+1}, \theta_{l+1}) \leftarrow (\beta_k, \theta_i) $
\end{algorithmic}
\end{algorithm}

For scaling to problems of arbitrary complexity, we choose Adam~\cite{kingma2017adam} for computing the gradient update in Alg. \ref{alg:usc} and \ref{alg:usa} with a termination condition $\mathcal{C}:=\mathds{1}(|J^\Box_k -J^\Box_{k-1}| < 10^{-3}) \lor \mathds{1}(J^\Box_k > J^\Box_{k-1} + 10^{-3}) $ to terminate under convergence or divergence due to unbounded updates for the corresponding objective $J^\Box$. It is worth noting that while the maximum number of tuning iterations in the target system $L$ is expected to be less for practical reasons, the maximum number of epochs $K$ can be arbitrary as they are based on the simulated rollouts in the model. 

\section{Results}
\label{sec:results}



The proposed approach is demonstrated on pairs of $\{$system, controller$\}$ with increasing complexity, as shown in Table~\ref{table:stc}. The tuning procedure is first validated on $\{$cartpole, LQR$\}$ to compare the performance improvements and qualitative nature of our tuning algorithms. Increasing the complexity of the controller, we then test if the trends hold for $\{$cartpole, DNN$\}$. Finally, the scalability of the approach is put to the test on high-dimensional $\{$quadruped, DNN$\}$, $\{$biped, MPC$\}$, and $\{$biped, TO$\}$ showcasing the improvement in performance across both sim-to-sim and sim-to-real deployment. For the above control tasks, the performance objective is either to \emph{stabilize} the system about an equilibrium state or \emph{track} a reference state trajectory. Thus, we choose the following differentiable tuning loss $c(x_t,x_t^*) := \|x_t - x_t^*\|^2_2 $, where $x_t^*$ is the desired state at time $t$. For states that evolve in different geometric spaces (e.g., position in $\mathbb{R}^3$ and quaternions in $S^3$), we compute an unweighted sum of the individual cumulative norm squared errors. The tuning domain for all experiments is built on a JAX~\cite{jax2018github} eco-system with  MuJoCo XLA (MJX) \cite{todorov2012mujoco} for the differentiable simulator, Flax~\cite{flax2020github} for the DNN controllers, Optax \cite{deepmind2020jax} for Adam.
For all experiments, we limit to a \emph{single rollout} in the target system per iteration and maximum tuning iteration of $L=5$ in Alg. \ref{alg:offline} to demonstrate that the developed approach is applicable to systems of arbitrary complexity. 
We take $K=100$ and in Eq. \eqref{eq:comb_prob}, we set $w^\text{task}=w^\text{sysId}=1$ and refrain from setup-specific tuning as we aim for a robust tuning algorithm that works across the systems and controller without having its own hyperparameters that need tuning. The specific details and performance improvement are now discussed, followed by ablative comparisons.  

\subsection{Tuning Performance}
\subsubsection{Cartpole Stabilization} For $\{$cartpole, LQR$\}$, a simplified cart pole dynamics is linearized about the upright unstable equilibrium, and the corresponding infinite-horizon LQR controller is pre-computed. 
The target system is considered to be $30\%$ heavier than the model 
to highlight the contribution of co-tuning.
For $\{$cartpole, DNN$\}$, a $[32,32]$ Multi-Layer Perceptron (MLP) is trained using PPO in IsaacLab \cite{mittal2023orbit} based on PhysX's~\cite{physx2025} physics.
We introduce an extreme mass discrepancy of $300\%$, as the DNN controller was robust to the transfer between the domains of near-identical dynamics.
\begin{figure}
\vspace{0.2cm}
    \centering
    \includegraphics[width=0.42\textwidth]{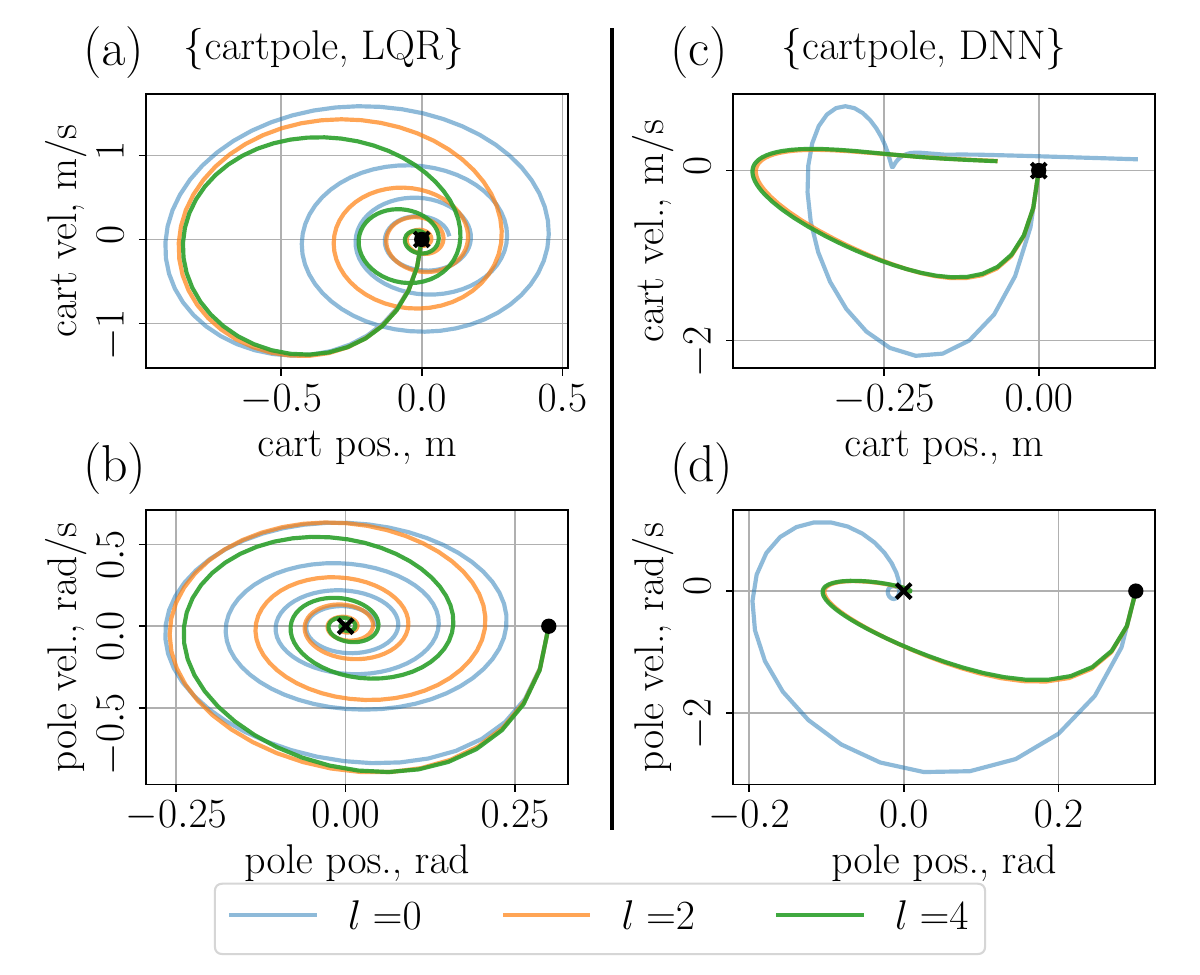}
    \caption{Phase portraits of progressive tuning iterates of DiffCoTune Split-Alternate on cart pole with model uncertainty. The initial and target states are marked by $\boldsymbol{\cdot}$ and $\boldsymbol{\times}$, respectively.}
    \label{fig:cp_trk_phase}
    \vspace{-0.8cm}
\end{figure}
The proposed DiffCoTune Split-alternate best improves the stabilizing performance for both LQR  and DNN controllers. For the LQR shown in Fig.~\ref{fig:cp_trk_phase}(a)(b), in successive tuning iterations, we see the phase portrait approaching a critically damped response (green) from the initial under-damped response (blue). 
The nominal DNN controller exhibits a steady-state velocity error (Fig. \ref{fig:cp_trk_phase}(c)) due to mass mismatch, causing drift and overshoot (blue), whereas tuning radically alters the phase portrait, yielding an over-damped response (green) that slowly converges to the target. In the pole's motion, Fig. \ref{fig:cp_trk_phase}(d), the tuned DNN controller gets critically damped with the shortest reach time. 
Notably, co-tuning led to an interpretable improvement for both controllers in the setpoint stabilization of a second-order system, from an underdamped, high-error response to a critically damped response with lower error.

\subsubsection{Quadruped Tracking} 
\begin{figure}
\vspace{0.2cm}
    \centering
    \centering
    \includegraphics[width=\linewidth]{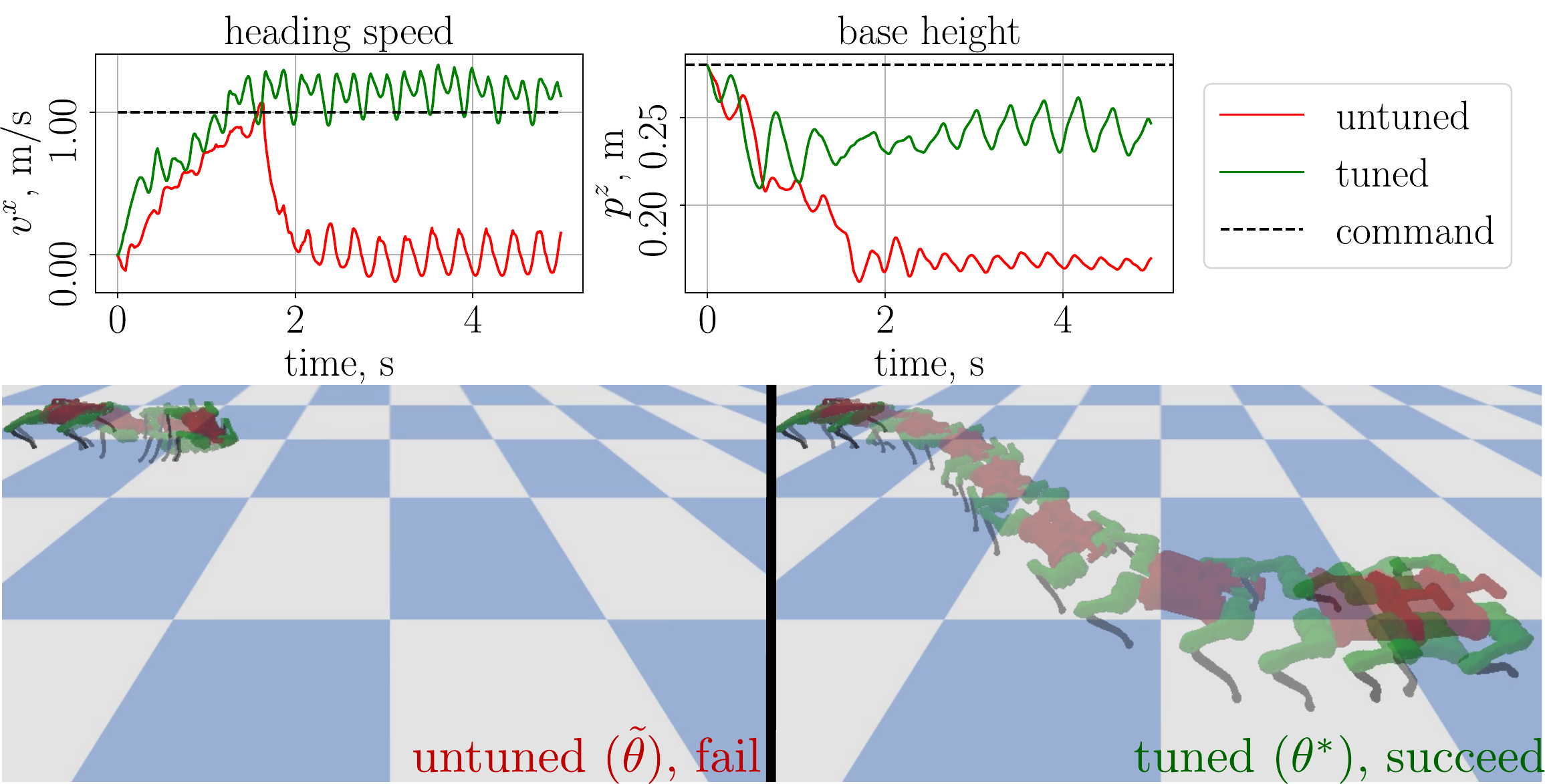}
    \caption{Sim-to-sim transfer of \emph{forward and yaw} motion of \{quadruped, DNN\} on the Barkour quadruped. }
    \label{fig:quad_f&y}
    \vspace{-0.2cm}
\end{figure}

To validate the scalability, we consider velocity tracking for quadruped locomotion both in sim-to-sim transfer with a Barkour robot \cite{caluwaerts2023barkour} and sim-to-real transfer with a Unitree Go1 robot, where a DNN controller ($[128, 128, 128, 128]$ MLP) is trained using PPO in BRAX \cite{brax2021github}. We directly tune the network's parameters, $\theta$, with dim($\theta$) = $112280$ and only consider the link masses for a tunable $\beta$. The tracking objective is defined with time-varying commands along heading velocity($v^x_t$), yaw rate ($\omega^z_t$) in the body frame with $v^y_t = 0$ to prevent lateral drift and $p^z_t = 0.28$ to maintain a nominal height. For sim-to-sim transfer, we consider a \emph{forward and yaw} motion: $v^x_t = 1.0$~m/s, $\omega^z_t = -0.7\sin(\frac{2\pi t}{T})$ rad/s  with Pybullet\cite{coumans2021pybullet} simulator as the deployment domain. With a LCP-based hard contact model contrasting the synthesis and tuning domain's soft contact model, a $15\%$ mass discrepancy is introduced in the deployment domain. The untuned controller falls immediately, Fig. \ref{fig:quad_f&y}, and fails to move any further subsequently. Tuning with DiffCoTune Split-Alternate remedies the fall and improves the performance by $60\%$ as seen in Fig. \ref{fig:quad_f&y}, with a near-perfect tracking of the desired velocities and the base height. 

\begin{figure}
    \centering
    \centering
    \includegraphics[width=\linewidth]{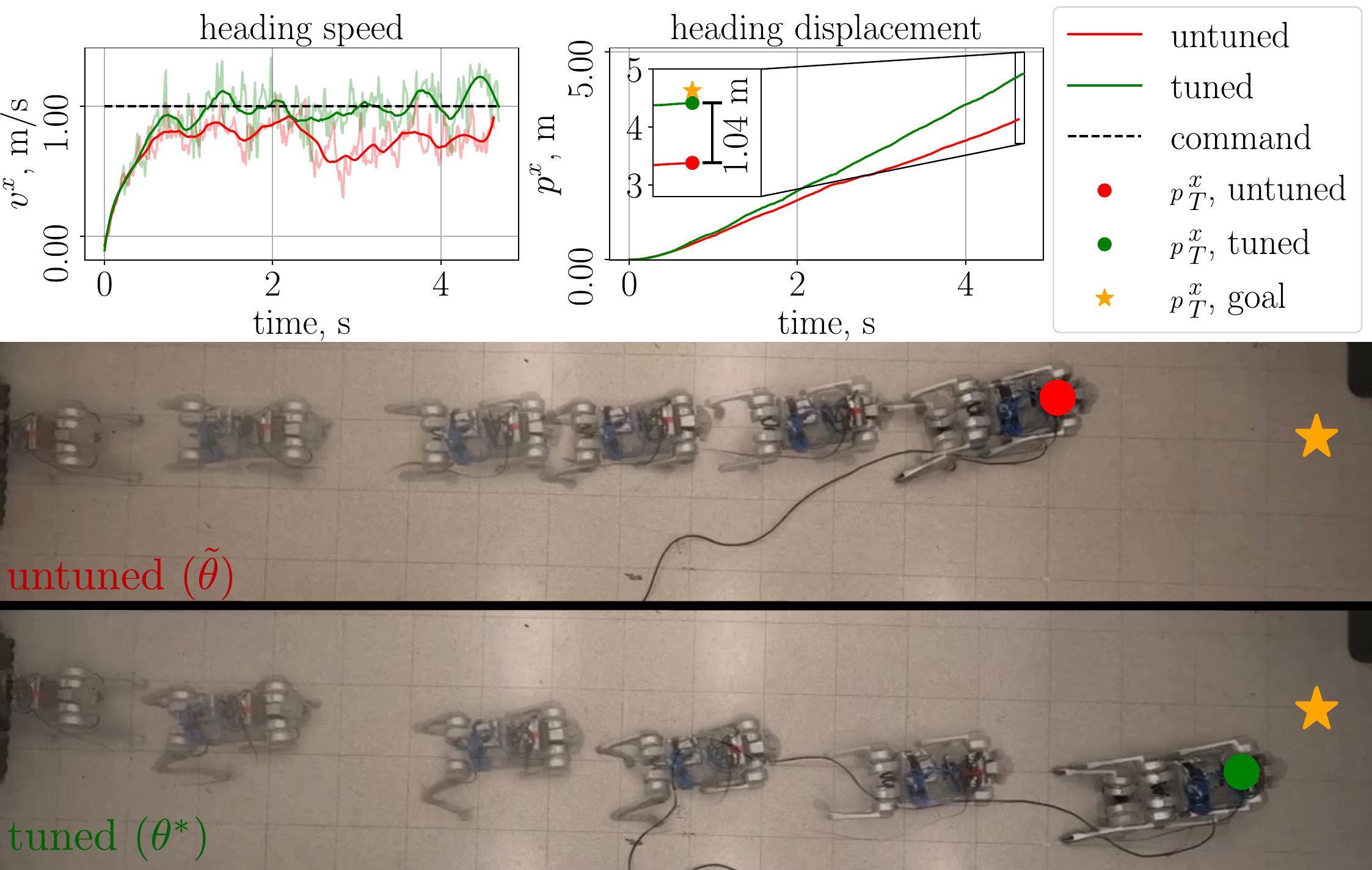}
    \caption{Sim-to-real transfer of \emph{high-speed forward} motion  of \{quadruped, DNN\} on a Unitree Go1 quadruped. }    
\label{fig:quad_hsf}
\vspace{-0.8cm}
\end{figure}

For sim-to-real transfer, we consider a \emph{high-speed forward} motion: $v^x_t = 1.0$ m/s with unknown model mismatches of the real hardware. While the untuned controller transfers successfully, it shows subpar tracking performance. In Fig. \ref{fig:quad_hsf}, filtering the noisy velocity feedback with a $5$-th order Savitzky-Golay filter, we observe an improvement in tracking through co-tuning. The improvement is more notable by visualizing the displacement, where the tuned controller advances $1.04$ m further than the untuned, reaching closest to the intended goal of $5$ m (tracking $1$ m/s for $5$ seconds), as also seen in the supplementary video.
\subsubsection{Biped Tracking}  Showcasing the flexibility of our approach, we tune optimal controllers formulated as Model Predictive Control (MPC) and Trajectory Optimization (TO) for sim-to-sim and sim-to-real transfer, respectively. 
We choose the MPC formulation for bipedal locomotion \cite{li2021force}, with a quadratic tracking objective and linearized Single Rigid Body Model (SRBM) dynamics, solved online as a Quadratic Program (QP). Through implicit differentiation~\cite{blondel2021efficient}, a QP's solutions are differentiable with respect to its parameters, thereby rendering it a differentiable controller. 
For the HECTOR~\cite{li2023dynamic} bipedal robot, we consider a \emph{height-tracking task} with desired base position commands in the world frame:  $p^z_t = 0.55 + 0.05\sin(\frac{2\pi t}{T})$, $p^x_t=0$, and $p^y_t=0$. Taking $\theta$ to be corresponding terms in diag($Q$) of the weight matrix $Q$ \cite{li2021force}, we add a clipping operator to our gradient update (like \cite{cheng2024diff}) to project $\theta$ into a feasible set ($Q \succeq 0$) to ensure the convexity of the QP.
As seen in the supplementary video, co-tuning improves tracking by $29.29$ \%, reducing jitters in the MuJoCo deployment domain despite the synthesis domain’s SRBM neglecting leg inertia, which is compensated by $\beta$ updates during tuning.
\begin{figure}[t!]
    \centering
    \includegraphics[clip, trim=0cm 0cm 12cm 0cm, width=1\columnwidth]{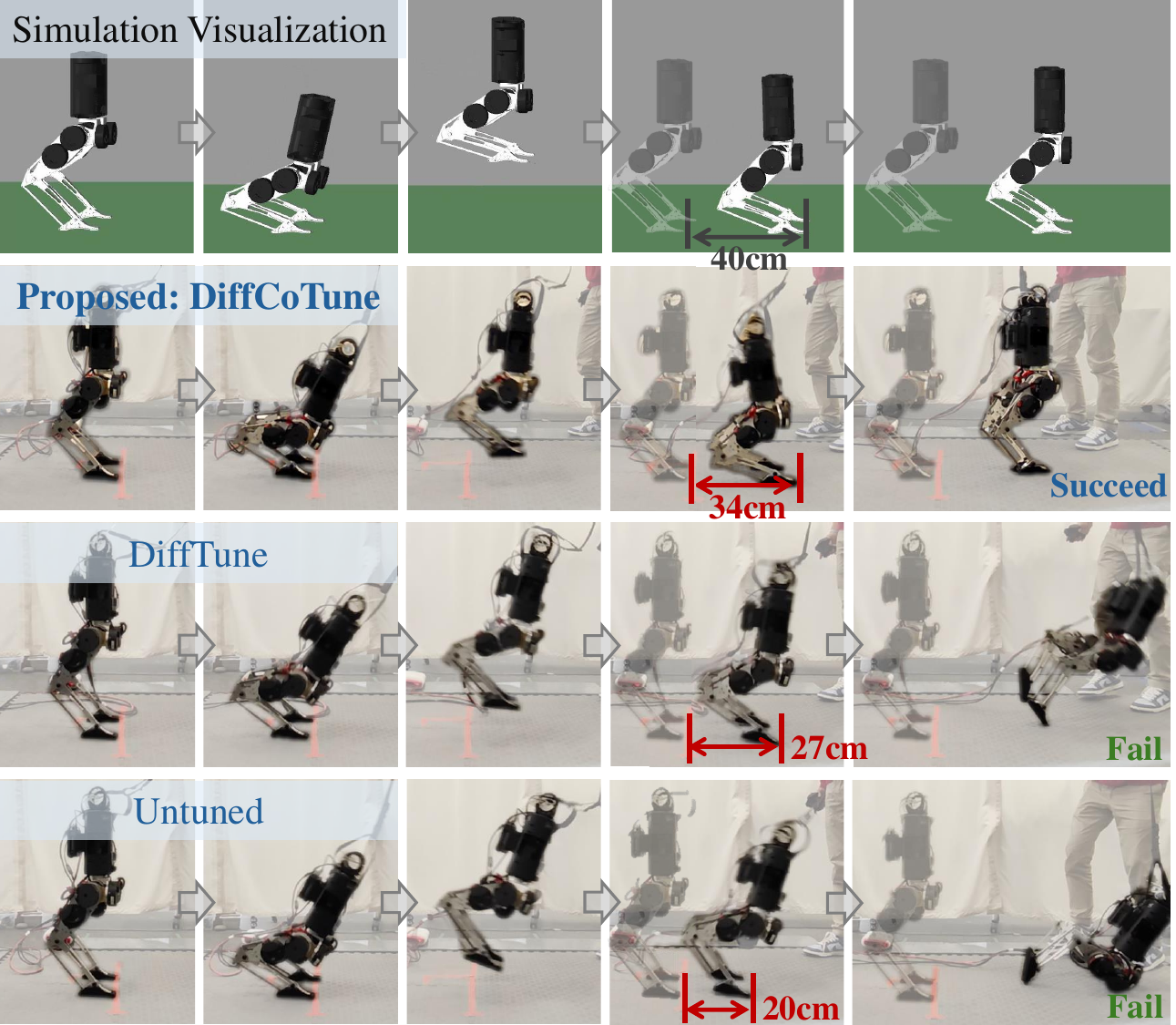}
    \caption{Comparative experimental snapshots of bipedal jumping with DiffCoTune, DiffTune, and unturned control parameters.}
    \label{fig:jumping}
    \vspace{-0.6cm}
\end{figure}

Secondly, we tune a dynamic bipedal jumping synthesized through TO \cite{li2024continuous}, solving a non-linear optimisation offline, subject to models of varying complexity to be transferred to the real robot. A tunable joint PD controller tracks the synthesized reference during takeoff and flight, parametrized with $\theta$ as time-varying $k_p$ gains. Having a single set of gains for both legs with $5$ doF each over a $100$ step long motion results in dim$(\theta)=500$.
 $\beta$ is set to link masses to account for off-board battery-induced mass uncertainty. As seen in Fig. \ref{fig:jumping}, an untuned controller fails to transfer to the real robot, while DiffTune improves jump length but falls upon touchdown. In contrast, DiffCoTune achieves a $34$ cm jump, closest to the synthesized $40$ cm jump, by (i) correcting torso pitch mid-flight and (ii) adjusting fore-aft foot placement for a stable landing as seen in Fig. \ref{fig:jumping} and the attached video, underscoring the transferability of our approach. 
 \subsection{Analysis and Comparisons}
We compare our Auto-Differentiation (AD) based approach with the relevant baselines and validate the need for co-tuning the model and control parameters through iterative rollouts. To study the effect of co-tuning, we compare with DiffTune~\cite{cheng2024diff} and Bayesian optimisation (BO) as gradient-based and gradient-free baselines, which only tune the control parameters. 
To highlight the significance of iterative data collection and tuning, we compare with a conventional first sysID and then control tuning as the corresponding baseline. Finally, we ablate the proposed approach over varying sources of uncertainty and design choices.
\subsubsection{Is Co-Tuning Necessary?}
For $\{$cart pole, LQR$\}$ without any model mismatches in Fig. \ref{fig:cp_sys_perf}(a), the AD variants of baseline DiffTune and proposed DiffCoTune perform similarly. With a mass discrepancy introduced in Fig. \ref{fig:cp_sys_perf}(b), tuning the controller parameters on the wrong model (DiffTune w/ model rollout) results in an initial improvement but plateaus, while DiffCoTune Split-Alternate outperforms in subsequent iterations. 
An identical trend is observed across $\{$cartpole, DNN$\}$ Fig. \ref{fig:cp_sys_perf}(c,d) and \{ quadruped, DNN \} Fig. \ref{fig:cp_sys_perf}(e,f).The improved performance of co-tuning is also qualitatively evident in $\{$biped, TO $\}$ as seen in the supplementary video and Fig. \ref{fig:jumping}
, highlighting the need for co-tuning.
\begin{figure}[t!]
    \centering
    \includegraphics[width=0.48\textwidth]{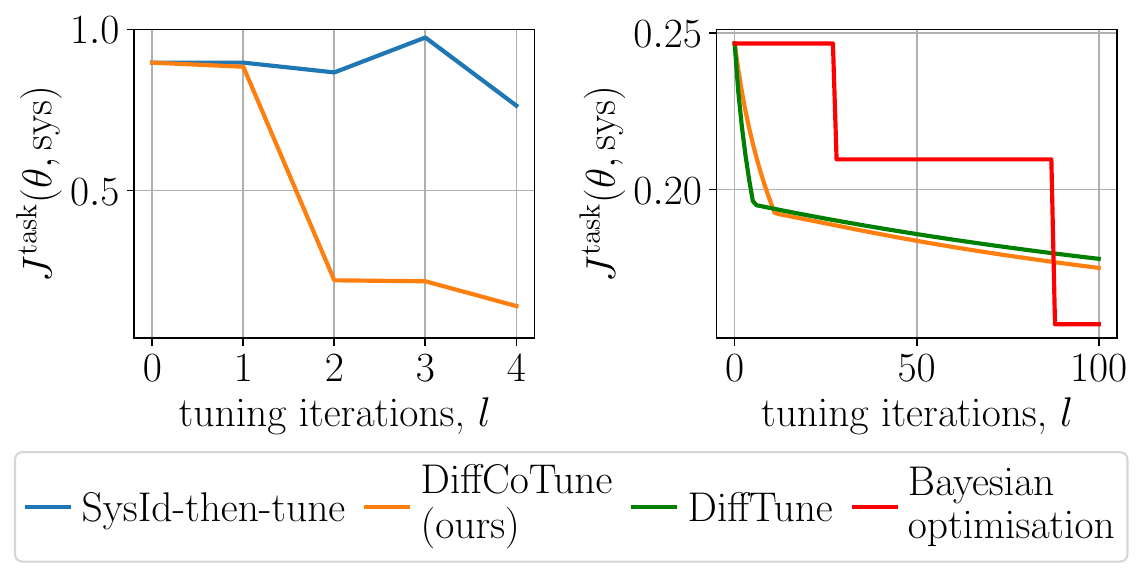}
    \caption{Performance loss comparison of the proposed DiffCoTune with \emph{sysId-then-tune} in \{quadruped, DNN\} (left) and  Bayesian optimization in \{cartpole, LQR\} }
    \label{fig:itr}
    \vspace{-0.8cm}
\end{figure}
\subsubsection{Gradient-based vs Gradient-free tuning} 
\begin{figure*}[h]
    \centering
    \includegraphics[width=\textwidth]{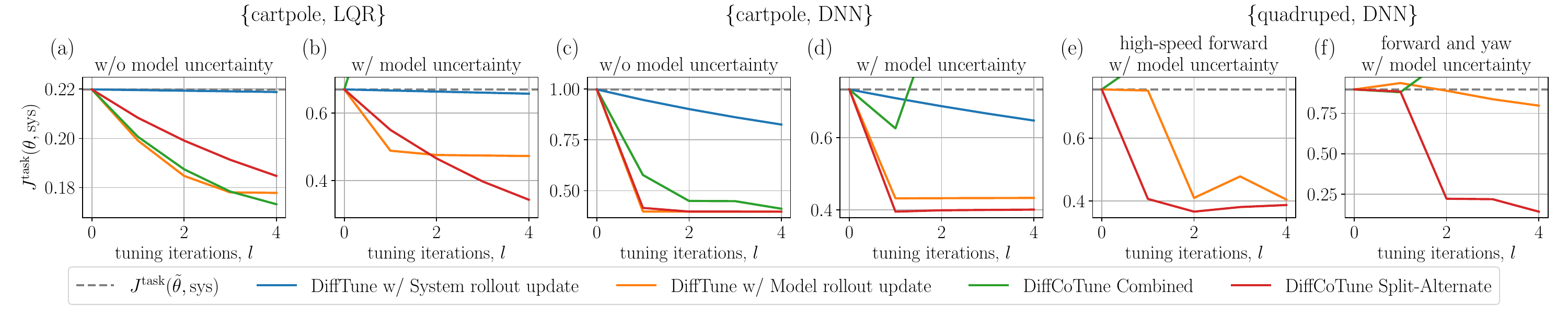}
    \caption{The performance loss along tuning iterations for tuning approaches over different $\{$system, controller$\}$ }
    \label{fig:cp_sys_perf}    
    \vspace{-0.6cm}
\end{figure*}

As a baseline for gradient-free tuning methods, we compare with Bayesian optimization (BO). For a fair comparison, we pick the \{cartpole, LQR\} as BO has not been shown to scale to high-dimensional parameter spaces ($>20$) \cite{berkenkamp2016safe} and tune for $100$ iterations in the target system for all methods, as shown in Fig. \ref{fig:itr} (right). We observe that the gradient-based DiffCoTune and DiffTune show a rapid reduction in the loss in and early transient phase ($l \leq 10$ trials), followed by a slower asymptotic decrease in the stationary phase ($l>10$).Notably, the rapid convergence of gradient-based methods persists even in high-dimensional tasks such as {quadruped, DNN}, Fig. \ref{fig:cp_sys_perf}(e,f). 
In contrast, being model-agnostic, vanilla BO requires more rollouts, ultimately finding the best solution after $90$ trials, unlike gradient-based methods that exploit first-order model information.
A rapid rate of improvement is crucial when target-system rollouts are costly, particularly for unstable systems and dynamic tasks like the showcased \{biped, TO\}. Moreover, BO required specifying tight bounds for the search domain of each parameter, which could be unintuitive for controllers parameterized by DNNs. For quantitative comparisons of gradient-based tuning with BO extensions and sampling-based methods, we direct the interested reader to prior work \cite{cheng2024diff}.
\subsubsection{Is Iterative Data Collection and Tuning Needed?}
A conventional sysId approach requires first tuning the parameters of a model and then either synthesizing or tuning the controller on the ``identified" model. In practice, for high-dimensional non-linear target systems, a nominal controller is rolled out from different initial conditions, commands, and perturbations to collect samples that hopefully excite parameters in the dynamic modes relevant to a given control task~\cite{jemin2019learning}. Similarly, we design the baseline: \emph{sysId-then-tune} by rolling out the untuned nominal controller from different initial conditions in the target system. In contrast, the proposed iterative scheme alternates between tuning the controller using collected data and collecting new data using the tuned controller (Alg. \ref{alg:offline}) analogous to \emph{on-policy learning} in RL. We set the number of target-system rollouts and tuning iterations to be the same for both approaches ($L=5$). 
While the baseline uses all the rollouts upfront for model update, followed by controller tuning, DiffCoTune iteratively collects data and co-tunes parameters.
The benefit of an iterative scheme was observed in the $\{$quadruped, DNN$\}$ setup as seen in Fig. \ref{fig:itr}(left). 
While the baseline improved performance after an initial deviation, DiffCoTune achieves greater loss reduction in the second step by using rollouts collected from the first tuned iterate instead of the initial nominal controller.
The supplementary video shows the baseline falling after a few steps, while our approach successfully completes the entire motion, validating the need for an iterative scheme.

\subsubsection{Controller Tuning with System Rollouts vs. Model Rollouts}
Inline with DiffTune \cite {cheng2024diff}, we consider a baseline propagating sensitivity using pre-computed model Jacobians along a target-system rollout (DiffTune w/ System Rollout Update), and a variant that solely tuned in a differential simulator purely with model rollouts (DiffTune w/ Model Rollout Update).
Assuming the target-system rollout lies in a close neighborhood of the model\footnote{which may be violated under model mismatch}, sensitivity propagation leverages parameter continuity to compute gradients~\cite{cheng2024diff} which can occur only once per system rollout ($K=1$, Alg.~\ref{alg:usc},\ref{alg:usa}). In contrast, DiffTune w/ model rollouts can perform any number of forward simulations and compute gradients($K \gg 1$) via AD, albeit not using target rollouts.
As shown in Fig. \ref{fig:cp_sys_perf}(a)--(d), while DiffTune w/ system rollouts (blue) improves performance, DiffTune w/ model rollouts (orange) significantly outperforms due to the higher number of gradient updates. Thus, DiffCoTune Split-Alternate (red) not only benefits from multiple gradient updates but also leverages the target system's rollouts through $\beta$-tuning, enabling it to outperform DiffTune w/ model rollouts under model uncertainty, as seen in Fig. \ref{fig:cp_sys_perf}(b,d,e,f).

\subsubsection{Combined vs. Split-Alternate}
Though the formulation of DiffCoTune Combined is straightforward, we find it to be fragile in practice in the presence of model uncertainty. For both LQR and DNN controllers, in the absence of model uncertainty, DiffCoTune Combined performs comparably to the better-performing variants as seen in Fig. \ref{fig:cp_sys_perf}(a, c). Upon introducing a model discrepancy,  DiffCoTune Combined consistently diverges for both cartpole and quadruped variants Fig. \ref{fig:cp_sys_perf}(b,d,e,f). Our naive choice of relative weights ($w^\text{task}=w^\text{sysId}=1$ in Eq. \eqref{alg:usa}) has exposed the sensitivity of the combined strategy to the curvature of the objective landscape, and we suspect this as the cause of its algorithmic fragility.
On the other hand, DiffCoTune Split-Alternate without any weight vectors, is immune to the diverse objective landscapes generated by varying systems and controllers of different structures across dimensions and scales as seen in Fig. \ref{fig:cp_sys_perf}(a)--(f), making it the scalable and robust tuning algorithm we set out to design.

\subsubsection{Varying source and magnitude of uncertainty}
\begin{figure}
    \centering
    \includegraphics[width=1\columnwidth]{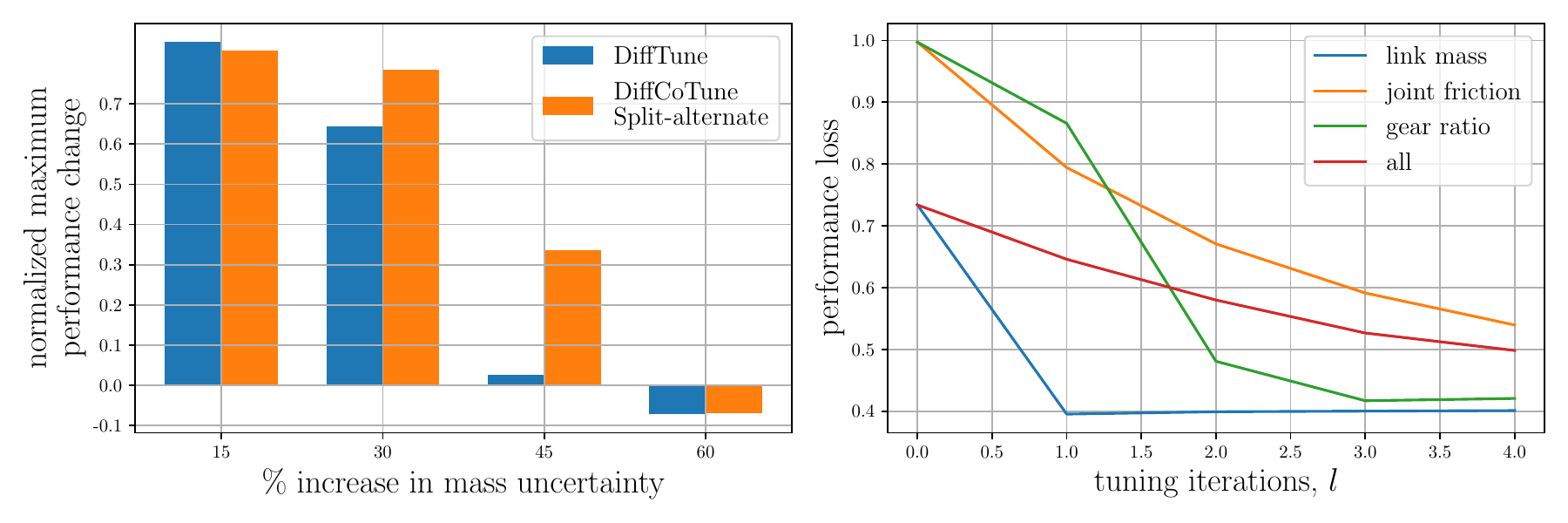}
    \caption{Effect of varying magnitude (left) and nature (right) of uncertainty}
    \label{fig:vuc}
    \vspace{-0.8cm}
\end{figure}

Under a moderate intensity of model mismatch, we note that DiffTune performs adequately (Fig. \ref{fig:cp_sys_perf}). Hence,  in \{quadruped, DNN\} setting, we compare the methods with increasing mass uncertainty. As seen in Fig. \ref{fig:vuc} (left), both methods show comparable performance improvement at $15\%$ mismatch, while at  $30\%$ and $45\%$, DiffCoTune outperforms DiffTune. At an extreme $60\%$ mismatch, both degrade performance as the nominal controller becomes invalid, shifting the problem from tuning back to synthesis. Additionally, to validate applicability under other mismatches, like friction and motor strength, we extend \{cartpole, DNN\} results from link mass (dim($\beta$) = 2) to tune for joint damping (dim($\beta$) = 2) and gear ratio (dim($\beta$) = 1). In Fig. \ref{fig:vuc} (right), DiffCoTune Split-alternate consistently improves performance across uncertainties and notably even on an amalgamation of all mismatches (dim($\beta$) = 5), highlighting its robustness.

\section{Conclusion}
\label{sec:conclusion}

In this work, we address the challenge of transferring a controller from its synthesis domain to the deployment domain by formulating it as a tuning problem. Specifically, we propose iterative co-tuning strategies to simultaneously tune the parameters of a differentiable controller and differentiable simulator to improve performance in the deployment domain with model uncertainty. Our approach only assumes that the model or controller is numerically differentiable, i.e., having numerical gradients or sub-gradients. Empirical validation was conducted on underactuated and hybrid systems like cart pole, biped, and quadruped, where we tuned distinct controller parameterizations synthesized using techniques such as LQR, MPC, TO and Deep RL. Experiments highlight the scalability of our approach through successful sim-to-sim and sim-to-real transfer. Future work will include extensions for tuning controllers over their entire region of attraction and co-tuning the DNN-based residual dynamics model to capture unknown unmodelled effects. 


\bibliographystyle{ieeetr}
\bibliography{references}

\end{document}